\documentclass[letterpaper, 10 pt, conference]{ieeeconf}  

\IEEEoverridecommandlockouts                              

\usepackage{graphics} 
\usepackage{epsfig} 
\usepackage{amsmath} 
\usepackage{amssymb}  

\usepackage{graphicx}
\usepackage{caption,subcaption}
\usepackage{wrapfig}
\usepackage{float}
\usepackage{cleveref}
\usepackage{multirow}
\usepackage{xcolor}

\usepackage[switch]{lineno}

\usepackage{placeins}

\title{\LARGE \bf
Autonomous Search for Sparsely Distributed Visual Phenomena through Environmental Context Modeling}

\author{Eric Chen$^{1}$, Travis Manderson$^{2}$, Nare Karapetyan$^{3}$, Peter Edmunds$^{4}$, Nicholas Roy$^{2}$, Yogesh Girdhar$^{3}$
\thanks{\scriptsize $^{1}$Massachusetts Institute of Technology and Woods Hole
Oceanographic Institution Joint Program
        {\tt\scriptsize ericrc@mit.edu}}%
\thanks{\scriptsize  $^{2}$Massachusetts Institute of Technology
{\tt\scriptsize \{travislm,nickroy\}@mit.edu}}%
\thanks{\scriptsize  $^{3}$Woods Hole Oceanographic Institution {\tt\scriptsize \{nare,ygirdhar\}@whoi.edu}
}%
\thanks{\scriptsize $^{4}$California State University {\tt\scriptsize peter.edmunds@csun.edu}
}%
}

\begin{document}

\maketitle
\thispagestyle{empty}
\pagestyle{empty}

\begin{abstract}
Autonomous underwater vehicles (AUVs) are increasingly used to survey coral reefs, yet efficiently locating specific coral species of interest remains difficult: target species are often sparsely distributed across the reef, and an AUV with limited battery life cannot afford to search everywhere. When detections of the target itself are too sparse to provide directional guidance, the robot benefits from an additional signal to decide where to look next. We propose using the visual environmental context -- the habitat features that tend to co-occur with a target species -- as that signal. Because context features are spatially denser and often vary more smoothly than target detections, we hypothesize that a reward function targeted at broader environmental context will enable adaptive planners to make better decisions on where to go next, even in regions where no target has yet been observed. Starting from a single labeled image, our method uses patch-level DINOv2 embeddings to perform one-shot detections of both the target species and its surrounding context online. We validate our approach using real imagery collected by an AUV at two reef sites in St. John, U.S. Virgin Islands, simulating the robot's motion offline. Our results demonstrate that one-shot detection combined with adaptive context modeling enables efficient autonomous surveying, sampling up to 75$\%$ of the target in roughly half the time required by exhaustive coverage when the target is sparsely distributed, and outperforming search strategies that only use target detections.
\end{abstract}



\section{INTRODUCTION}
Coral reefs are ecologically critical ecosystems that support a large range of biodiversity, and monitoring these ecosystems is essential for understanding their health and informing conservation efforts. A key scientific task is to capture images of as many instances as possible of a target coral species, enabling researchers to study the distributions, abundances, and overall health of that species.


\begin{figure}[ht]
\centering
\includegraphics[width=0.95\columnwidth]{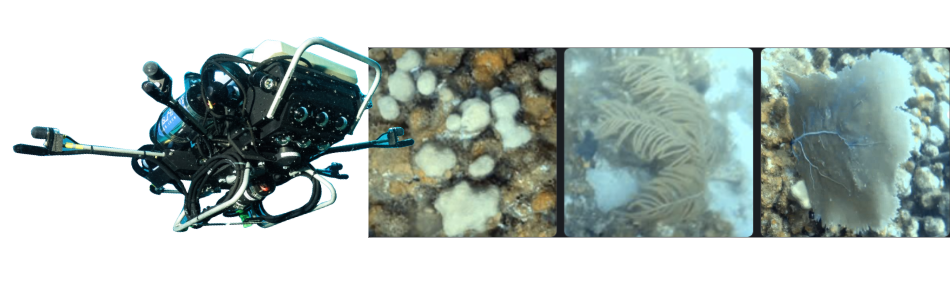}
\vspace{-2ex} 
\par 
\begin{minipage}{0.5\columnwidth}
  \centering
\includegraphics[width=0.95\textwidth]{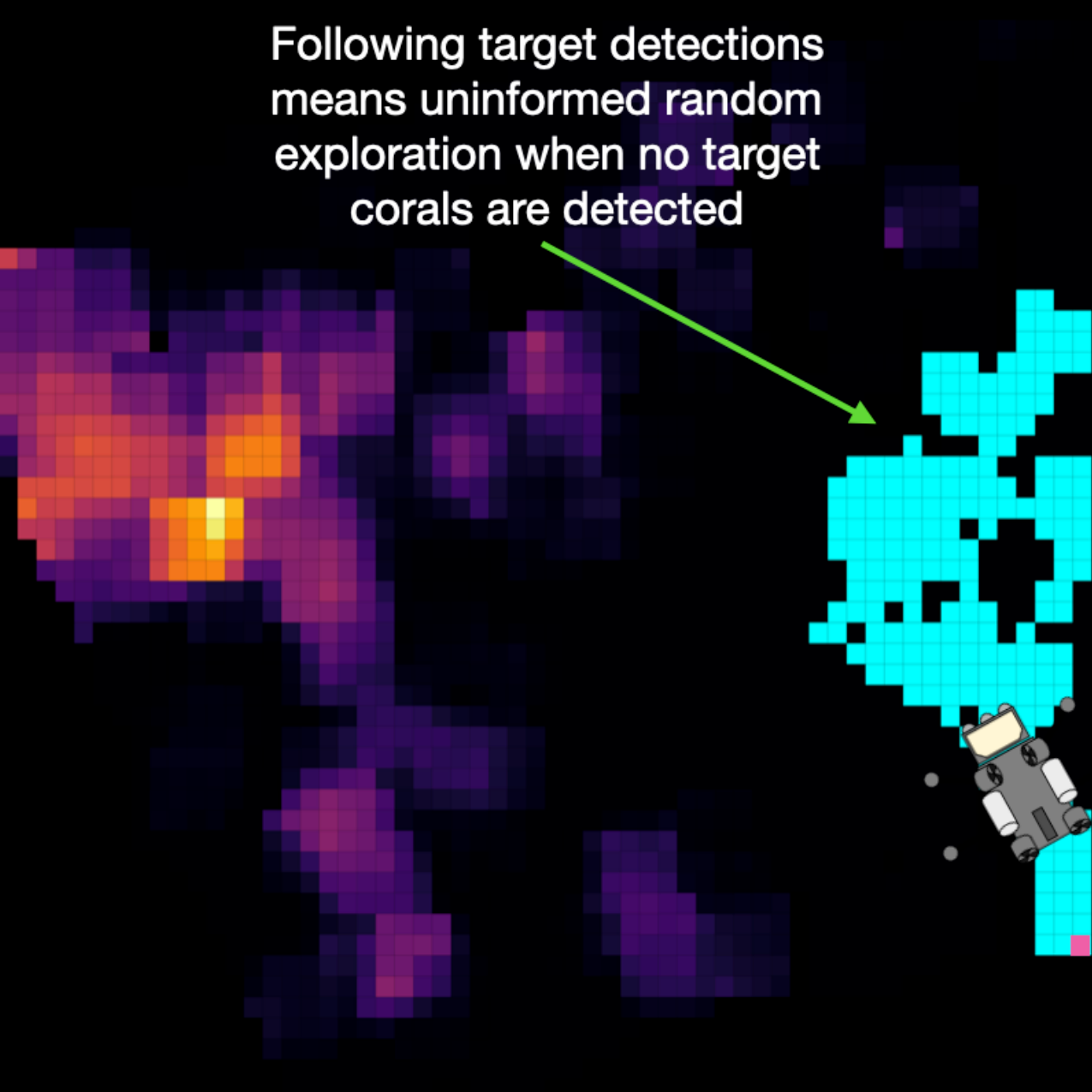}
\subcaption[.]{Target Species}\label{fig/paths/target}
\end{minipage}%
\begin{minipage}{0.5\columnwidth}
  \centering
\includegraphics[width=0.95\textwidth]{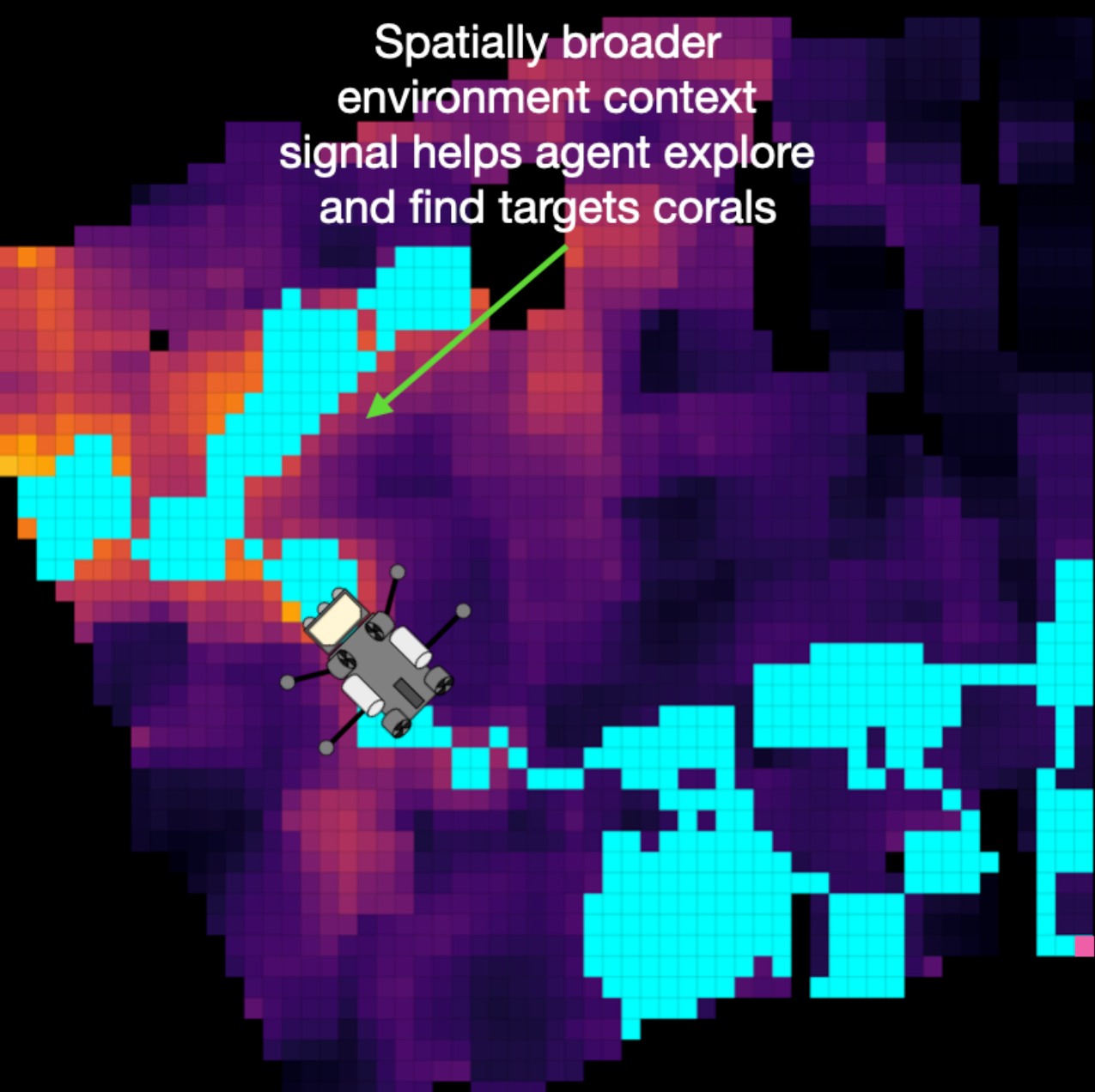}
\subcaption[.]{Environment Context}\label{fig/paths/ec}
\end{minipage}
\caption{Example of the AUV surveying for one of the three coral species used to evaluate our approach (depicted above). Two same-length robot trajectories (blue squares) generated by (a) following target species detections vs. (b) environment context detections. The heatmap of environment context detections (b) guides the robot towards regions with t arget species even in areas where the target species heatmap (a) contains no information and forces random exploration.}
\label{fig/paths}
\vspace{-3ex}
\end{figure}


Surveying reefs manually with divers or teleoperated vehicles is resource intensive and limited in spatial coverage. Autonomous underwater vehicles (AUVs) offer a scalable alternative to manual diver surveys, but traditional fixed-pattern surveys such as "lawnmower" patterns do not adapt to the distribution of a target species, often resulting in sub-optimal sampling efficiency. Adaptive surveying approaches that tailor AUV search behavior to individual reef site characteristics and different target species distributions can improve AUV efficiency, collecting larger numbers of scientifically valuable samples.

However, adaptive surveying of specific species on coral reefs poses a fundamental challenge. Target coral species are often sparsely distributed across a reef: even with a reliable detector, positive detections are rare and scattered, providing the robot with no directional signal for where to find more of the target. Without additional guidance, the robot must fall back on random actions, wasting limited battery life over large regions where the target is absent (Figure \ref{fig/paths}).

This search problem can be formalized as computing a distribution $Q = p(c \,|\, \mathbf{x})$, where $c$ is the target coral class and $\mathbf{x}$ denotes spatial coordinates. Given a robot path $R = {\mathbf{x}_1, \ldots, \mathbf{x}_N}$, we can estimate $p(c \,|\, \mathbf{x})$ from collected observations. The straightforward approach is to choose $R$ as a coverage path, but this is slow. A greedy approach that selects waypoints to maximize $p(c \,|\, \mathbf{x})$ directly fails when the target coral distribution is patchy and sparse, leaving the robot with no gradient to follow between isolated detections. If the robot can detect environmental context alongside the target itself, it gains a spatially broader exploration signal that provides directional guidance even in regions where no target has yet been observed (Figure \ref{fig/paths}).

We hypothesize that different coral species occupy distinct environmental sub-niches within a reef at spatial scales of meters to tens of meters. We refer to these co-occurring habitat features as the species' \textit{environmental context}, and we propose that $R$ can be improved by modeling this environmental context $E$ such that $p(c \,|\, \mathbf{x}) \propto p(c \,|\, E)p(E \,|\, \mathbf{x})$, and then greedily computing $R$ to maximize $p(E \,|\, \mathbf{x})$. While a target species may be sparse, the broader habitat in which it occurs --- substrate textures, neighboring organisms, and reef structures that tend to surround it --- is typically denser and varies more smoothly across the reef. 

We operationalize detections using patch-level feature embeddings from DINOv2 \cite{oquab2023dinov2}, a self-supervised visual foundation model. While off-the-shelf segmentation models are rapidly advancing \cite{kirillov2023segment, sauder2025coralscapes, zheng2024coralscop}, we found empirically that they often do not reliably detect the full range of coral species considered in this work, particularly smaller species in visually complex reef environments. Feature correspondences in the DINOv2 embedding space enable one-shot detection of these species from a single labeled image, without any retraining. To characterize environmental context, we sample non-target features that co-occur with the target and maintain them in an online buffer that adapts as the robot explores new areas of the reef. Our entire approach is one-shot: only a single target image needs to be labeled by selecting a few patches corresponding to the target coral species.


We evaluate this framework by simulating robot trajectories using real reef imagery by an AUV at two sites in St. John, USVI, showing that environment context-guided exploration consistently improves survey performance across both sites and three biologically different coral species (Figure \ref{fig/paths}).
  
\textbf{The main contributions of this paper are:}

\begin{itemize}
\item \textbf{Demonstration of one-shot detections of three diverse coral species across two reef sites on real AUV imagery collected in St. John, USVI.}
\item \textbf{A method for online characterization of the visual environment context of a target coral species that enables more efficient target-conditioned surveying.}
\end{itemize}

\section{METHOD}
We consider the problem of target-conditioned surveying in coral reef environments, where the objective is to maximize the number of observations of a specified target coral species within a fixed time horizon. The survey platform is an AUV equipped with a downward-facing camera, operating at an approximately fixed altitude above the reef and collecting image observations as it traverses the site. The environment is treated as approximately planar, allowing the AUV’s trajectory to be restricted to a two-dimensional survey domain. At each step, the vehicle must decide where to move next given the current image observation in order to maximize its cumulative reward, defined as the number of target species detections collected within the available mission time.

\begin{figure*}[ht]
\centering
\begin{minipage}{0.3\textwidth}
  \centering
\includegraphics[width=0.85\textwidth]{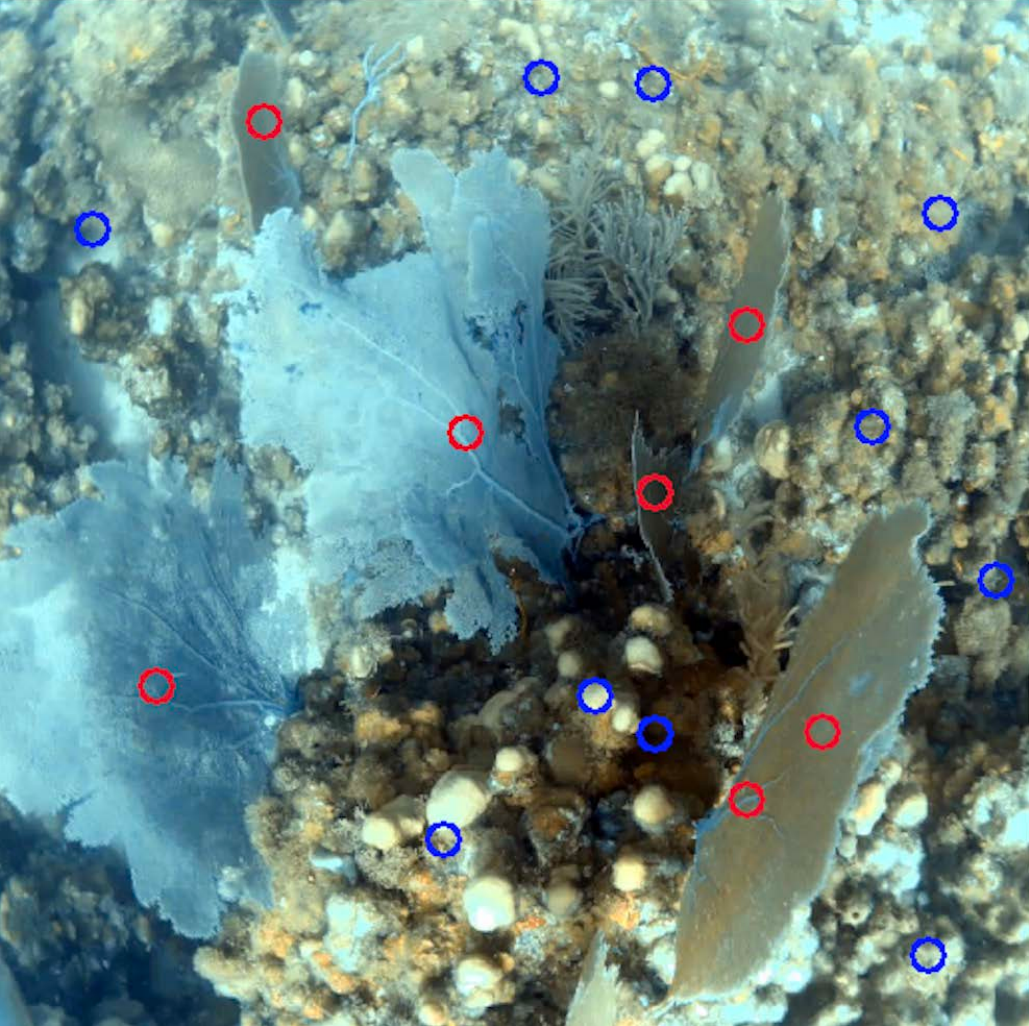}
\subcaption[first caption.]{Target Image}\label{fig/detections/query}
\end{minipage}%
\begin{minipage}{0.3\textwidth}
  \centering
\includegraphics[width=0.85\textwidth]{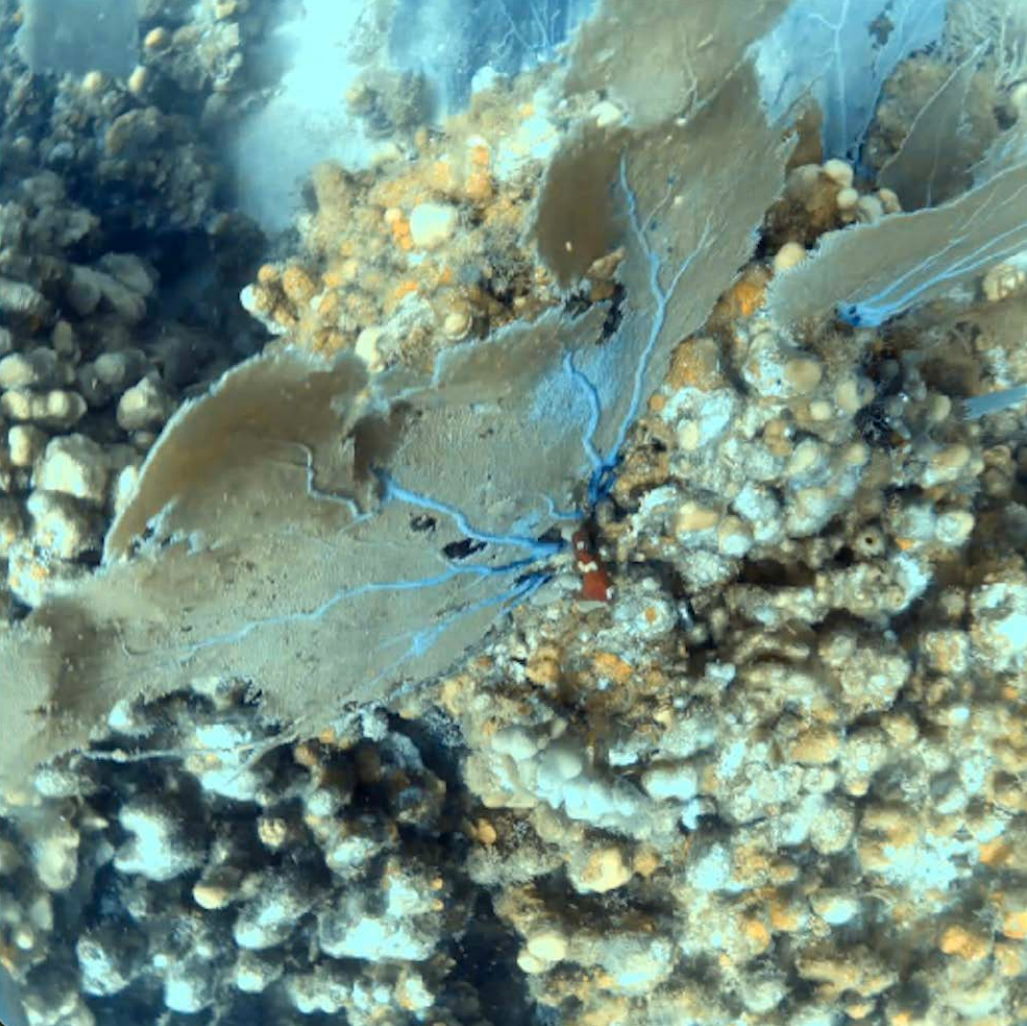}
\subcaption[second caption.]{Candidate Image}\label{fig/detections/candidate}
\end{minipage}%
\begin{minipage}{0.3\textwidth}
  \centering
\includegraphics[width=0.85\textwidth]{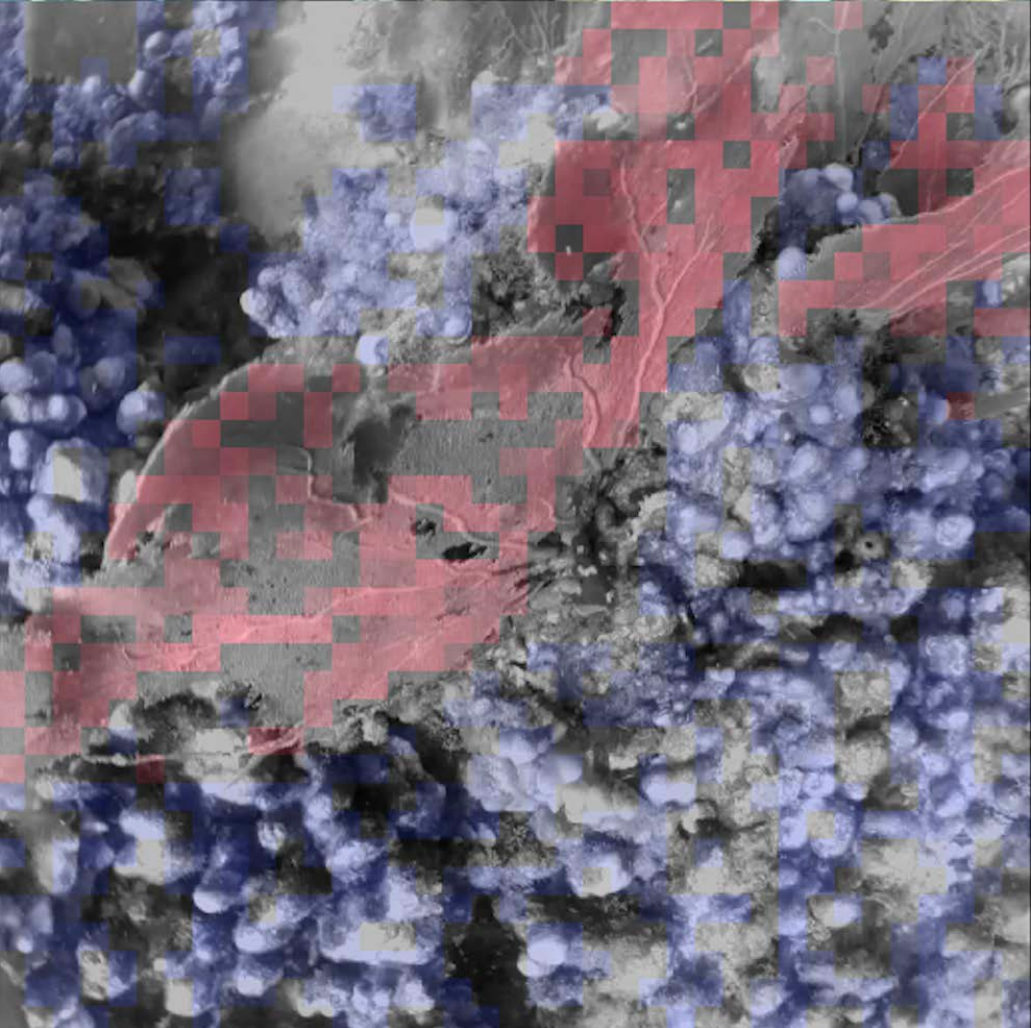}
\subcaption[third caption.]{Class Detections}\label{fig/detections/class}
\end{minipage}

\caption{Given selected patches from a target image (a), a candidate image (b) can be labeled with class assignments computed from patch-level correspondences (c). Red circles mark a target species (\textit{Gorgonia ventalina}), and randomly sampled blue circles characterize the initial environment context model for this species. Patch-level class assignments (c) show that our approach detects the target coral species (red patches) and surrounding environment context (blue patches).}
\label{fig/detections}
\end{figure*}

For an operator deploying our methodology, the workflow proceeds as follows: an AUV is deployed at a site of interest and teleoperated until the operator identifies a target species. The operator then labels a handful of image patches corresponding to the target (Figure \ref{fig/detections}). These patches initialize the one-shot detector, while the environment context model is simultaneously initialized from randomly sampled non-target patches in the same image. The vehicle then transitions to autonomous operation, detecting new target instances, updating its context model online, and using both signals to guide exploration. Our approach consists of three components: (1) one-shot detection of target coral species, (2) image-level scoring of species presence, and (3) environment context detection.


\subsection{One-Shot Detector}

We first aim to detect candidate patches in an unlabeled survey image that correspond to a given target species, given a single labeled image. Formally, given an input image $I \in \mathbb{R}^{H \times W \times 3}$, DINOv2 partitions the image into non-overlapping consecutive patches of size $P \times P$ pixels to obtain the set,
$$
\mathcal{P}(I) = \{ p_1, p_2, \ldots, p_N \}, \quad N = \frac{H}{P} \times \frac{W}{P}.
\eqno{(1)}
$$

Each image is center-cropped such that $H=W=518$, and $P=14$ (default input resolution and patch size for the DINOv2 ViT base model) such that $|\mathcal{P}(I)| = 1369$. After the image is partitioned into patches $\mathcal{P}(I)$, these patches are passed through a pretrained DINOv2 vision transformer $f_\theta(\cdot)$, which outputs a set $\mathcal{X}(I)$ of $d$-dimensional embeddings corresponding to the feature embeddings for each patch $p_i$,
$$
\mathcal{X}(I) = f_\theta(\mathcal{P}(I)) = \{ x_1, x_2, \ldots, x_N \}, \quad x_i \in \mathbb{R}^d. \eqno{(2)}
$$

To perform detection for a target class $c$, we assume the existence of a single target image $I_q$, from which a set of $M$ target feature embeddings $Q_c \subset \mathcal{X}(I_q)$ are generated from target patches,
$$
Q_c = \{ q_1, q_2, \ldots, q_{M} \}, \quad q_j = f_{\theta}(p_j) \in \mathbb{R}^d.
\eqno{(3)}
$$

Practically, a human operator selects these features by clicking on a few patches in the target image that correspond to the target coral species of interest. For each new candidate image patch embedding $x_i = f_{\theta}(p_i)$ and each target feature embedding $q_j$, we compute the cosine similarity,
$$
s_{i,j} = \frac{\langle x_i, q_j \rangle}{\|x_i\| \, \|q_j\|}.
\eqno{(4)}
$$

We define the correspondence score of patch $p_i$ with class $c$ as the maximum similarity across target feature embeddings,
$$
S_i^c = \max_{j} s_{i,j}.
\eqno{(5)}
$$

To eliminate weak correspondences, we threshold these scores,
$$
\tilde{S}_i^c = 
\begin{cases}
S_i^c & \text{if } S_i^c \geq \sigma_c. \\
0 & \text{otherwise.}
\end{cases}
\eqno{(6)}
$$

Thresholds were selected by inspecting cosine similarity distributions on held-out images from the first reef site and held fixed for all subsequent experiments. A target threshold of $\sigma_c = 0.3$ eliminates noisy detections, while a context threshold of $\sigma_c = 0.1$ admits a broad range of features without including weak correspondences. Finally, to ensure that each patch is assigned at most one class (to eventually avoid conflating the environment context with the target), we apply a tiebreaking rule, assigning each patch to the class with the maximum nonzero score,
$$
\hat{c}_i = \arg\max_c \tilde{S}_i^c, \quad \hat{S}_i = \max_c \tilde{S}_i^c.
\eqno{(7)}
$$

\subsection{Image-Level Scoring}
While our patch-level detection produces per-patch correspondences, our goal is to obtain an image-level score that reflects the strength of presence for a target species. Since patches may only partially cover object contours and are not guaranteed to align precisely with species instances, we adopt a heuristic that counts the number of patches per-image assigned to each class. For a target class $c$, we define the image-level score as the sum of the number of patches in that image that match a target patch,
$$
\Phi_c(I) = \sum_{i=1}^N \mathbf{1}[\hat{c}_i = c].
\eqno{(8)}
$$

The patch-counting heuristic provides a proxy for species abundance and coverage within the image. While segmentation would yield precise instance masks, patch-level features are coarse and this approach provides a robust, empirically validated estimate. The assumption that the AUV maintains a roughly constant altitude ensures that $\Phi_c(I)$ correlates with the actual spatial coverage of the target species.

\subsection{Environment Context}
Beyond detecting target species, we can also characterize the visual environmental context of the target, under the assumption that certain visual features tend to co-occur spatially with each species. Because the target coral species is sparse, individual positive or negative detections from an image rarely provide clear guidance on which direction to move next. In contrast, the environmental context signal is typically denser across an image, offering a spatially denser signal that suggests promising directions of exploration. The planner leverages this signal to guide movement in ways that preserve or increase exposure to relevant environmental context.

For a target class $c$, we define an environment context feature set $\mathcal{E}_c$, which contains representative patch embeddings not assigned to $c$ but frequently observed in the vicinity of class $c$.

To initialize $\mathcal{E}_c$, we leverage the target image $I_q$, which is guaranteed to contain at least one instance of the target class. We first identify the complement set of background patches in image $I_q$ that do not correspond to the target class $c$,
$$
\mathcal{X}_{\bar{c}}(I_q) = \{ x_i \in \mathcal{X}(I_q) \;|\; \hat{c}_i \neq c \}.
\eqno{(9)}
$$

From here, we randomly sample $K$ features from this set of non-target patches $\mathcal{X}_{\bar{c}}(I_q)$ and insert them into a context buffer, subject to a maximum buffer size $M$,
$$
\mathcal{E}_c \leftarrow \text{Sample}_K\big(\mathcal{X}_{\bar{c}}(I_q)\big), \quad |\mathcal{E}_c| \leq M.
\eqno{(10)}
$$

\noindent
where $I_q$ denotes a labeled target image containing target class $c$, and $K$ and $M$ are chosen hyperparameters. Sampling this way ensures that $\mathcal{E}_c$ is initialized with background features from regions that are spatially adjacent to true instances of the target.

\begin{figure}[h]
    \centering
    \includegraphics[width=\columnwidth]{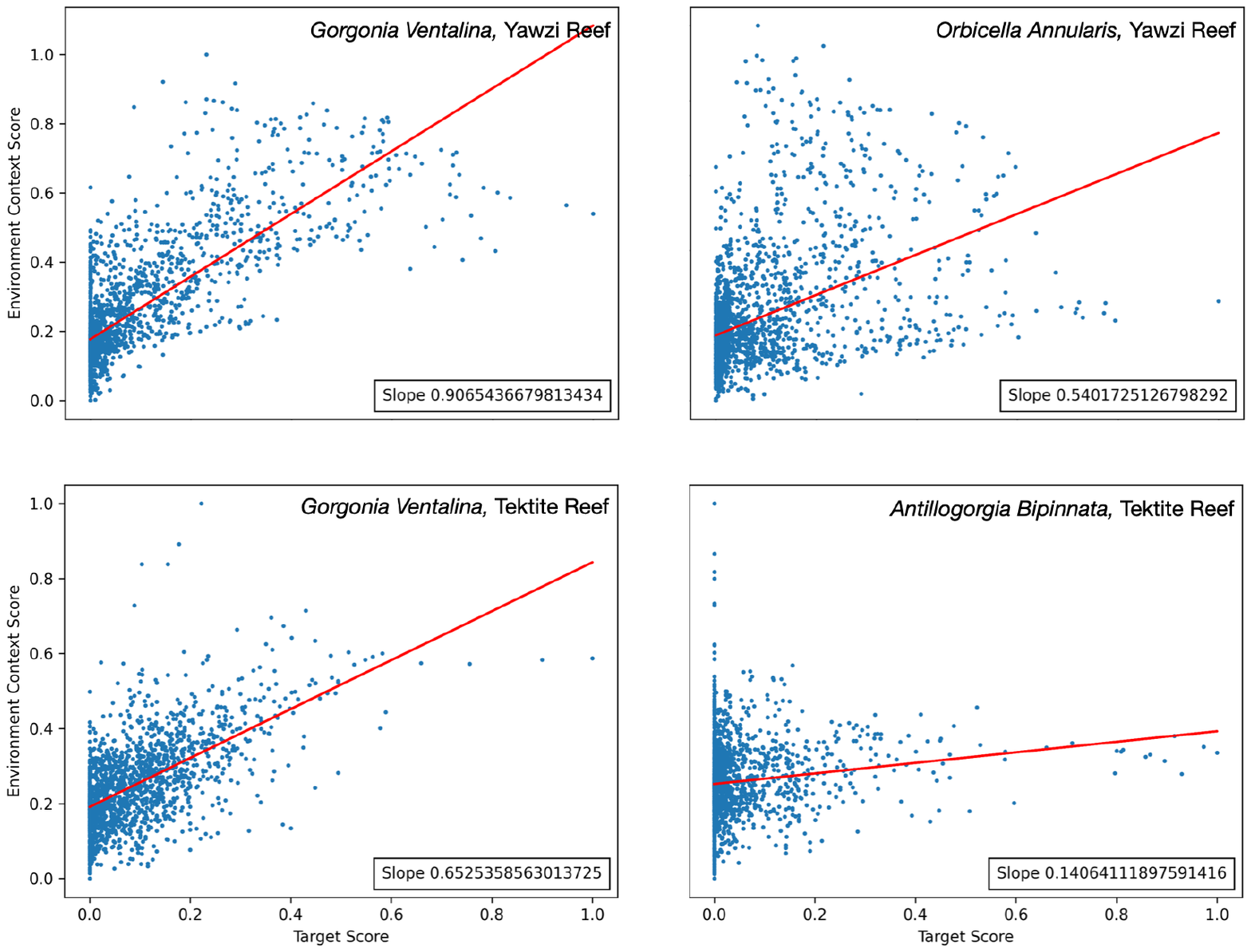}
    \caption{Comparison of normalized target species detection scores and target environment context detection scores across spatial cells in both reef site orthomosaics. The positive regression slopes indicate a consistent spatial co-occurrence between target and context scores. While the correlation is modest, the context signal nonetheless provides useful directional information for the greedy policy, which only needs to compare a small set of neighboring cells rather than make precise abundance predictions.}
    \label{fig/monotonicity}
\end{figure}

In order to capture the variance in the environment context of a target species across an entire reef site, the context buffer $\mathcal{E}_c$ is updated by adding $K$ randomly sampled background features from any image $I$ that scores strongly enough for target species presence. We find that for the size of the reef sites considered in our evaluation (20m x 20m), catastrophic forgetting is avoided by selecting $M=200$, and $K=25$. For larger sites with greater context variance, further parameter tuning or alternative buffer update strategies may be needed to avoid catastrophic forgetting. Specifically, for each image $I$, we compute the image-level score $\Phi_c(I)$ (Eq. 8), and if this exceeds a user-defined threshold $\tau_c$, we update the buffer by sampling new features from $\mathcal{X}_{\bar{c}}(I)$,
$$
\text{if } \Phi_c(I) \geq \tau_c:
\mathcal{E}_c \leftarrow \text{FIFO}_M\Big(\mathcal{E}_c, \text{Sample}_K(\mathcal{X}_{\bar{c}}(I))\Big),
\eqno{(11)}
$$

\noindent
where $\text{FIFO}_M(\cdot)$ denotes a bounded first-in, first-out queue of maximum size $M$. If adding new samples would cause $\mathcal{E}_c$ to exceed $M$, the oldest features are discarded, ensuring that the buffer always maintains a fixed memory footprint while adapting dynamically to new observations and potential spatial drift in the environment context. We find that our approach generates context scores that align with target presence (Figure \ref{fig/monotonicity}), and that using a context buffer that is updated online outperforms a context buffer that is fixed at initialization (see results).

\section{EXPERIMENTAL EVALUATION}

\begin{figure*}[ht]
    \centering
    \includegraphics[width=0.9\textwidth]{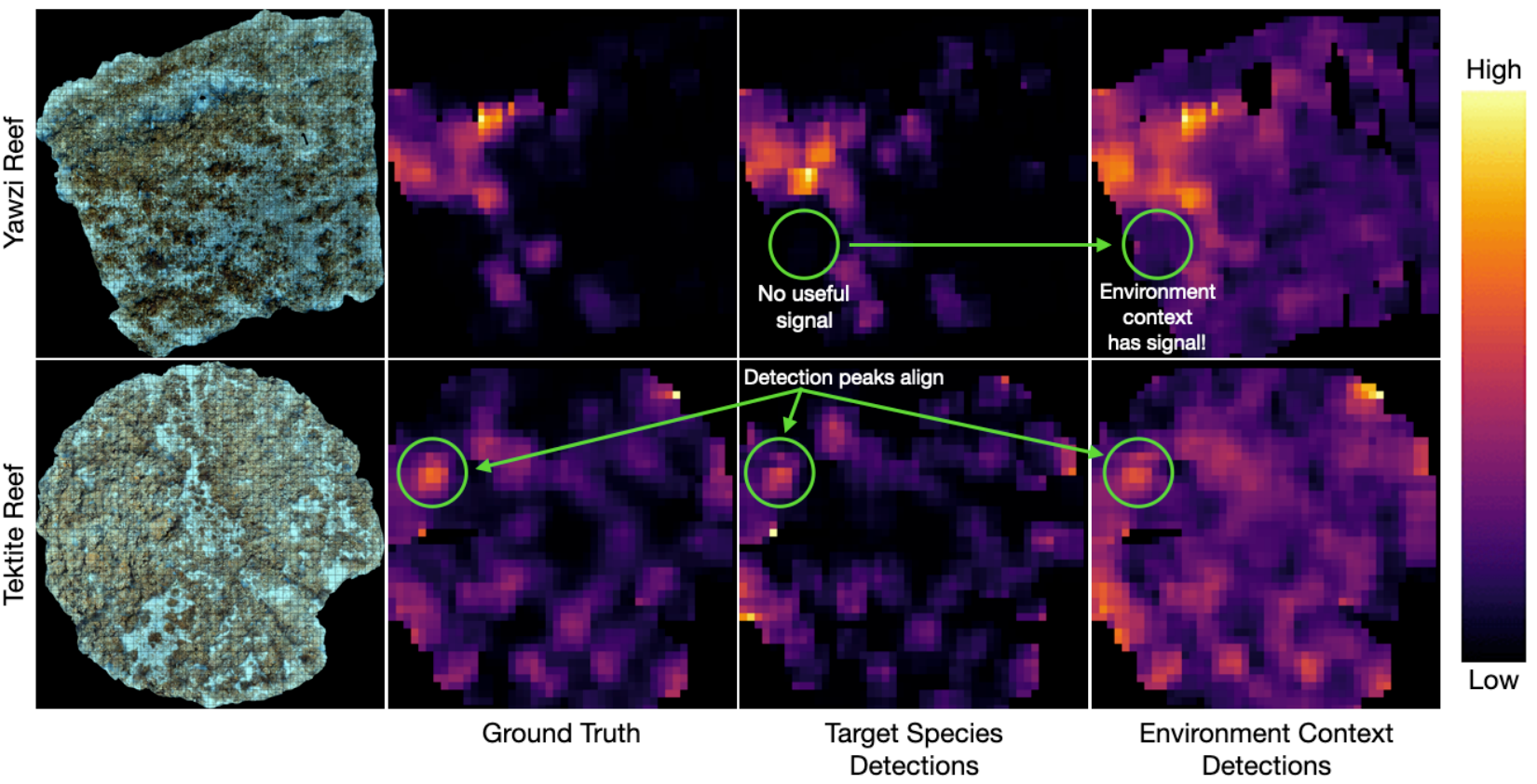}
    \caption{Ground truth (hand-labeled) and detected distributions for \textit{Gorgonia ventalina} at Yawzi Point and Tektite reefs in St. John, USVI. The peaks in ground truth, target species, and environment context detections align, although the environment context distribution has planning signal that leads towards target peaks, in areas where the target distribution has none.}
    \label{fig/orthos}
\end{figure*}

To evaluate the utility of environment context detections in generating target-specific survey patterns, we test our approach on real imagery collected in March 2024 by an AUV at two 20m × 20m reef sites (Yawzi Point and Tektite) in St. John, USVI. We evaluate our approaches ability to survey three target coral species (Figure \ref{fig/paths}): \textit{Gorgonia ventalina}, \textit{Orbicella annularis}, and \textit{Antillogorgia} spp.. These three coral species were chosen due to their visual and biological differences. \textit{Orbicella annularis} is a stony coral that is often smaller in size than \textit{Gorgonia ventalina} and \textit{Antillogorgia} spp., is attached to the sea floor (i.e., sessile), and engages in biological processes that build stony reef structure. \textit{Gorgonia ventalina} and \textit{Antillogorgia} spp. are both soft corals that sway with seawater movement and do not build reef structure. The variance in visual appearance for soft species like \textit{Gorgonia ventalina} and \textit{Antillogorgia} spp. makes them interesting targets for the one-shot DINOv2 detector. Additionally, the fact that these corals are biologically dissimilar means that the environment context for each of them is likely quite different, making them good targets to evaluate the generality of our environment context generation approach.

Approximately 4,000 images were processed in MetaShape \cite{metashape} to construct orthomosaics of each site (Figure \ref{fig/orthos}). The orthomosaic serves not as synthetic input, but as a spatial reference frame that positions each real AUV image within the site. Each image is registered to the orthomosaic using its center pixel coordinates and footprint, allowing us to associate grid cells with the set of real images that overlap them spatially. The alignment of the images to a coordinate frame enables us to retrieve representative observations for any grid cell, so that arbitrary robot trajectories can be replayed on real reef imagery. In effect, the orthomosaic provides the spatial scaffold for simulating survey strategies, while ensuring that every evaluation is grounded in actual visual data collected by the AUV.

\subsection{Context-driven Exploration}
We compare two exploration strategies: a coverage policy and a greedy myopic policy guided by local observations. Fixed survey patterns such as lawnmowers and straight lines have been standard for many practitioners conducting underwater surveys \cite{hagen2003hugin, jakuba2024exploring}. In this work, the coverage baseline (\textbf{Lawnmower}) is implemented as a lawnmower\cite{choset2000coverage} trajectory with row spacing equal to one grid cell width, ensuring every cell is visited exactly once. 

In contrast, the greedy myopic policy adapts its path based on local image observations. The survey domain is discretized into a regular grid of cells, where each 3×3 block of cells corresponds to approximately one full AUV downward-facing camera footprint (Figure \ref{fig/orthos}). At each time step, the vehicle has access to the image crops that cover the cell it occupies and the eight surrounding cells. The candidate image is the image corresponding to the central grid cell in the 3×3 grid cell region. Each of the eight surrounding cells are scored for target species presence and environmental context using the DINOv2-based approach described above, and the vehicle moves to the neighboring cell with the highest score. To encourage exploration and prevent cycling, the score of any visited cell is zeroed upon first entry; if all surrounding cells score zero, the vehicle selects a random unvisited neighbor.

\subsection{Scoring}
The greedy myopic policies start from random positions in each trial, and all policies are run for the same number of time steps. \textbf{The maximum number of timesteps corresponds to the length of the full coverage lawnmower, which visits every cell once and exhaustively samples all of the target coral species present at the reef site}. To approximate coral colony number and size observed, each trajectory is scored using the pixel area of ground truth segmentation masks for each of the three target coral species. All curves using a local greedy policy represent an aggregated performance across 100 random trials.

\subsection{Baselines}
We compare the performance of each policy when using: 1) target detections only (\textbf{Target}), 2) target detections and target environment context detections (\textbf{Target + EC}), and 3) target environment context detections only (\textbf{EC}) as an exploration signal to select which cell to visit next. 

We also compare a range of alternative exploratory signals against our target environment context detections. While the co-occurrence relationships of different coral species is still an area of active research in the marine biology community, it is generally accepted that corals tend to grow on rocky substrata rather than bare sandy bottoms. In order to demonstrate that our algorithmically generated notion of target environment context does not reduce to a simple substrata detection algorithm, we trained a segmentation model to detect substrata. This model consists of a trained multilayer perceptron (MLP) head on top of a frozen backbone \cite{sauder2025coralscapes}. In order to compare against a manual definition of environment context as substrata, we evaluate the performance of a greedy myopic policy that uses substrata segmentations only (\textbf{Segmentation}), and substrata segmentations and target detections (\textbf{Segmentation + Target}) as an exploration signal.

\section{RESULTS AND DISCUSSION}

\begin{figure*}[ht]
    \centering
    \includegraphics[width=\textwidth]{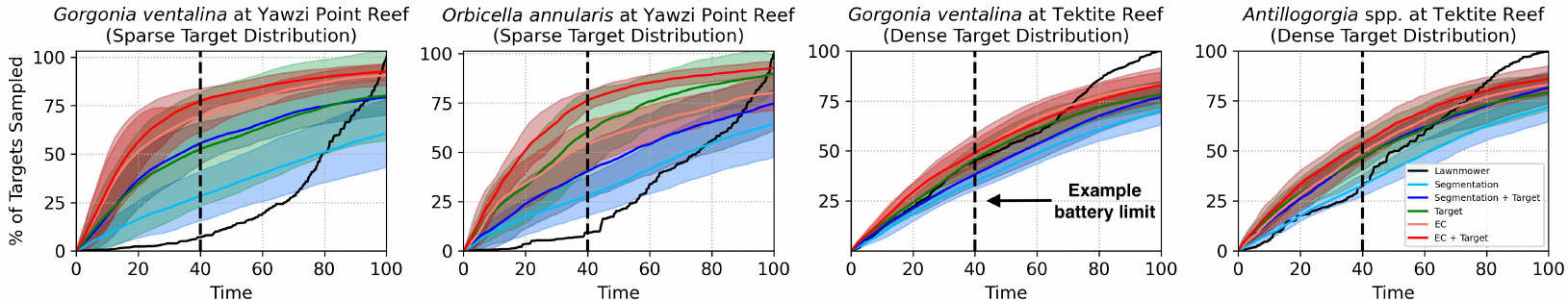}
    \caption{Percent of all possible targets sampled against normalized time steps for different surveying policies (exhaustive lawnmower and greedy local search) for each target coral species at both reef sites. Shaded regions denote $\pm$1 standard deviation across 100 random seed trials of the greedy policy. At full coverage for each site (Time = 100), the lawnmower samples 100$\%$ of the target. Our method significantly outperforms a naive lawnmower pattern, sampling 75$\%$ of the target vs. 10$\%$ at Time = 40 when targets are sparsely distributed. These test sites are spatially bounded (20m x 20m) --- in larger environments where exhaustive coverage is infeasible, the rapid initial collection rate suggests a greedy policy that leverages environment context could potentially provide practical advantages.}
    \label{fig/reward}
    \vspace{-3ex}
\end{figure*}

\subsection{Environment Context}

We first evaluate the effectiveness of combining target detections and environment context in guiding AUV surveys. Figure \ref{fig/reward} shows reward curves that represent the total number of instances of the target coral sampled as a function of time. In each plot the maximum number of timesteps corresponds to exhaustive coverage for the coverage policy (lawnmower), which fully samples the target species. Across both reef sites and multiple target species, following the combined signal of environment context and target detections consistently results in better performance than all other methods, including using target detections as an exploration signal alone. This result emphasizes the spatial breadth of the environment context signal: the environment context signal provides useful directional information in regions where target detections alone provide no directional information which allows the vehicle to make informed decisions instead of taking random actions, and to perform better than following target detections alone.

Importantly, the greedy policy that follows the combined signal of environment context and target detections samples more target species substantially faster than the greedy policies that follow all other signals. 75$\%$ of all the \textit{Gorgonia ventalina} at Yawzi Point reef is sampled in 50$\%$ less time than the exhaustive coverage lawnmower. A fast convergence rate is a highly desirable feature for practitioners, since marine surveys can often be interrupted prematurely in the field due to changing environment conditions or unforeseen technical difficulties with the vehicle. While the sites covered here are bounded (20m x 20m), the rapid initial target collection rate suggests a greedy policy that leverages environment context could provide practical advantages at larger sites where exhaustive coverage is infeasible.

These results also show that our environment context characterized using deep visual features creates a better exploratory signal than manually defining environment context by segmenting for substrata. Following only the environment context performs similarly with following target detections, whereas following only substrata segmentations underperforms all methods. This result suggests that our notion of environment context aligns better with the presence of target species than substrata segmentations.


Interestingly, the performance differences between all methods are more pronounced for the Yawzi Point reef site than they are for the Tektite reef site. We hypothesize that the large performance gap at Yawzi Point is due to the difference in the ground truth distribution of the target coral species at each of these sites. As seen in Figure \ref{fig/orthos}, the distribution of \textit{Gorgonia ventalina} at Yawzi Point reef is much more isolated to the left side of the site, meaning that exploration becomes crucial if starting anywhere on the right side of the site. The distribution of the same species at Tektite reef however, is much more uniformly spread out across the entire reef site. As a result, regardless of where the mission begins the AUV will already be spatially close to target coral species, and the difference between random exploration and exploration using environment context is reduced. While our method generally outperforms all others for both species at each site, it is important to note that at the Tektite reef the coverage policy lawnmower achieves a higher number of total instances sampled at full coverage. While our method converges more rapidly, the multi-modal nature of the distribution of target coral species at Tektite reef means that the local myopic policy can easily be distracted by local maxima, allowing the coverage policy to outperform the greedy policy in the long run. 


\begin{figure*}[ht]
    \centering
    \includegraphics[width=\textwidth]{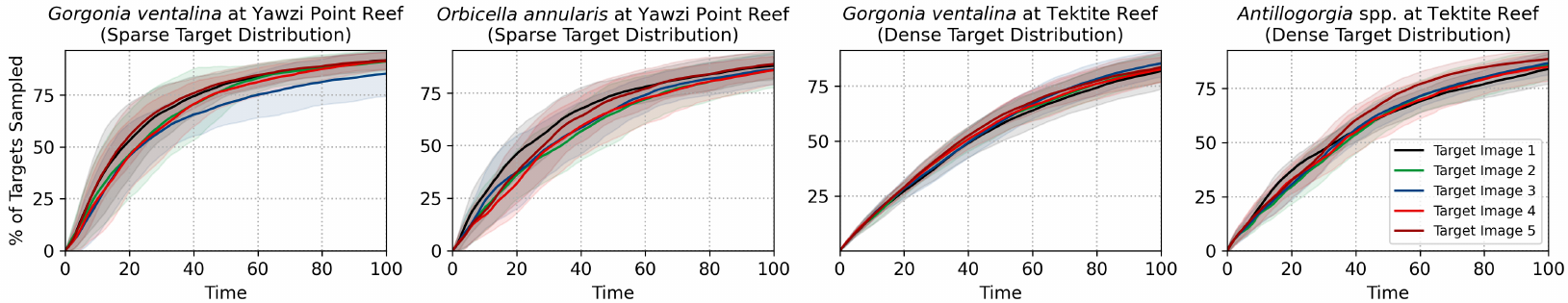}
    \caption{Percent of all possible targets sampled for the \textbf{EC + Target} policy following environment context signals generated from five different initial target images. Shaded regions denote $\pm$1 standard deviation across 100 random seed trials of the greedy policy. There are slight differences between performance at the Yawzi Point site, but overall our method is fairly insensitive to variations in the initial target image.}
    \label{fig/qtest}
    \vspace{-1ex}
\end{figure*}

\begin{figure*}[h]
    \centering
    \includegraphics[width=\textwidth]{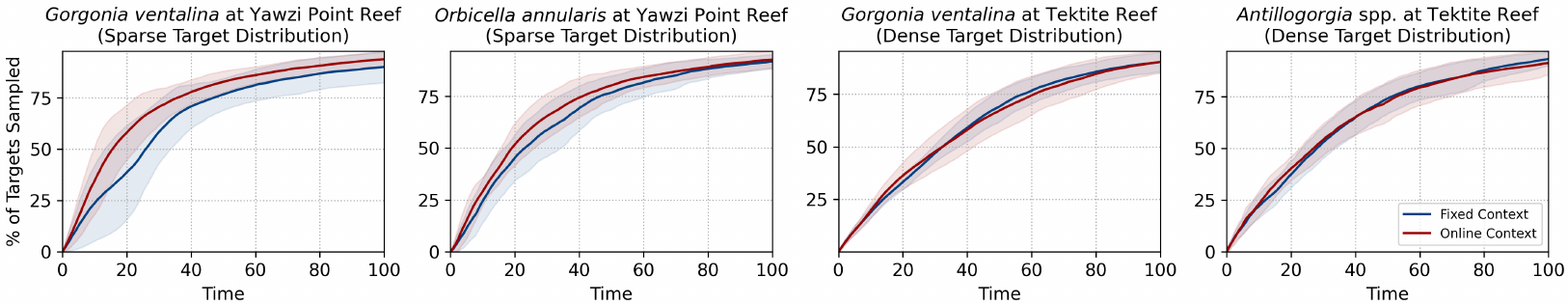}
    \caption{Percent of all possible targets sampled for a policy that follows the signals generated by an environment context buffer that is fixed at initialization (\textbf{Fixed Context}), and an environment context buffer that is updated each time a target species is observed (\textbf{Running Context}). Shaded regions denote $\pm$1 standard deviation across 100 random seed trials of the greedy policy. In environments with higher environment context variance (Yawzi Point), updating the environment context buffer outperforms the environment context buffer that is fixed at initialization.}
    \label{fig/fixrun}
    \vspace{-2ex}
\end{figure*}

\subsection{Sensitivity to Initial Target Image}
We also evaluated the sensitivity of our method to the choice of initial target image. For each species and site, we randomly selected five images and provided target exemplars by clicking on instances of the target within each image. The full pipeline was then run using these initializations, and we report reward curves in Figure \ref{fig/qtest} that show performance aggregated over 100 random seeds for a greedy local-search policy that follows the combined environment context and target detections (\textbf{EC + Target}). While minor differences in performance are observed --- particularly at the Yawzi Point site --- overall the results indicate that our method is largely insensitive to the specific choice of initial target image. This robustness is important for practical deployment, as it suggests that operators in the field need not carefully search for an “ideal” first instance of a target coral species in order to obtain strong survey performance.

\subsection{Sensitivity to Online Buffer Updates}
We next evaluated the role of updating the environment context buffer online. Specifically, we compared two variants: a fixed buffer baseline (\textbf{Fixed Context}), where the environment context is populated once from features sampled in the initial target image and then held constant for the duration of the survey, and a running buffer baseline (\textbf{Running Context}), where the buffer is similarly initialized but subsequently updated whenever an image exceeds the target presence threshold. As shown in Figure \ref{fig/fixrun}, the performance of the two methods is largely indistinguishable at the Tektite site. In contrast, at Yawzi Point the running buffer exhibits a modest but consistent advantage over the fixed baseline. We hypothesize that this difference arises from the degree of environmental variability: the datasets considered here span relatively limited spatial extents (only 20m x 20m), which reduces the potential benefit of dynamically updating the buffer. Nonetheless, the results at Yawzi Point indicate that even within small reef sites, meaningful variation in species-specific environmental context exists, and that online updating can help capture this variation to improve survey efficiency. We expect online updating to potentially become more important at larger sites where visual environmental context varies more substantially, but confirming this would require experiments at larger spatial scales. That said, the strong performance of even the fixed buffer suggests that a single target image captures a substantial portion of the relevant environmental context at these spatial scales, which is a practical advantage for deployment --- the method does not depend on extensive online learning to be effective.

\section{RELATED WORK}

\subsection{Species Detection Underwater}
A large body of work has explored coral detection and segmentation for ecological surveys. Early methods relied on classical image processing \cite{beijbom2012automated}, while recent approaches use deep learning for semantic segmentation and object detection \cite{mahmood2017deep, gonzalez2020monitoring}. Large-scale efforts such as CoralNet \cite{chen2021new}, CoralScape \cite{sauder2025coralscapes}, and CoralSCOP \cite{zheng2024coralscop} introduced extensive labeled datasets and demonstrated the potential of supervised learning, but these methods can be difficult to generalize across sites with varying visual conditions. Since annotated imagery is costly to obtain, adapting such models to new locations or species remains challenging.

Recent advances in self-supervised learning and foundation models offer a path forward through one-shot detection. Models such as CLIP \cite{radford2021learning} and DINO/DINOv2 \cite{caron2021emerging, oquab2023dinov2} learn transferable feature embeddings and have shown promise in ecological applications \cite{docherty2024upsampling, amir2021deep, ayzenberg2024dinov2, ong2025evaluation, shao2025multi}. Hamilton et al. \cite{hamilton2022unsupervised} used self-supervised features for unsupervised segmentation, though not on underwater imagery, while Raine et al. \cite{raine2024human} leveraged DINOv2 features for coral species annotation but did not explore environmental context. Here, we demonstrate that pretrained patch-level embeddings can enable one-shot detection of both coral species and their associated visual contexts, using only a single labeled target image as supervision.

\subsection{Adaptive Surveying}
Autonomous underwater surveys have traditionally relied on methods that use prior environmental knowledge or build maps during deployment for planning \cite{flaspohler2019information, binney2010informative, pizarro2004large, berild2024efficient}. While effective, these approaches consume valuable mission time constructing environment representations before exploration can begin. Online surveying methods instead allow robots to adapt directly in novel sites, maximizing observation time. Girdhar et al. \cite{girdhar2014autonomous} introduced the Realtime Online Spatiotemporal Topic (ROST) framework, enabling robots to learn environmental features and adapt strategies on the fly, which has been applied to reef monitoring. However, such frameworks focus on general exploration rather than species-specific objectives.

Goal-conditioned approaches introduced by Karapetyan et al. \cite{karapetyan2021human,lin2024uivnav}, Manderson et al. \cite{manderson2020vision}, Todd et al. \cite{todd2024adaptive}, and Jamieson et al. \cite{jamieson2020active} adapt robot behavior for targeted exploration. Karapetyan et al. rely on trained segmentation models, limiting generalization. Manderson et al. propose a goal-conditioned visual policy that balances covering target species and reaching waypoints. Todd et al. characterize context using manually defined substrata classes from higher-altitude imagery. Jamieson et al. develop a reward model guided by ROST topics and human-provided exemplars. In contrast, our method requires only a single labeled target image at deployment, eliminating continuous human supervision while still enabling target-conditioned adaptive surveying.

Building on this line of work, we introduce a model that learns the environmental context of target species online using self-supervised features rather than pre-defined substrata classes. By capturing co-occurrence patterns between species and their surrounding habitat, our method provides a dense exploration signal that improves target-adaptive surveying in visually complex reef environments.

\section{CONCLUSION}
We presented a method for one-shot detection and context-aware adaptive surveying of coral reef environments using patch-level features from a pretrained vision transformer. Given a single labeled target image, our approach detects arbitrary coral species and learns the visual environmental context in which they occur, producing a spatially dense exploration signal that improves survey efficiency beyond following target detections alone. Results of simulated trajectories run on real reef imagery from surveys done in St. John, USVI, demonstrate that combining one-shot detection with environment context-guided reasoning offers a promising pathway toward efficient ecological monitoring. Several limitations of the current work suggest directions for future research. Our evaluation compares against a lawnmower baseline and detection-only variants on bounded 20m $\times$ 20m grids; stronger baselines such as information-driven planners \cite{todd2024adaptive, jamieson2020active} would better contextualize the benefit of the context model. The greedy policy also cannot cross large featureless gaps between reef patches, motivating extension to non-myopic planners. Deployment in larger, unbounded environments would further test generality --- particularly for the online context buffer --- and validating the full pipeline on robot hardware at larger, more diverse reef sites is an important next step.






\section*{ACKNOWLEDGMENTS}
Eric Chen is supported by an NDSEG Fellowship. This work was supported by NSF EAGER OCE 23-33604, ONR Uncertainty-based Active Learning N00014-22-1-2677, and NPS Permit \# VIIS-2025-SCI-0005.  Thanks to John Walsh, John Cast, Levi Cai, the Reef Solutions team, and the WHOI Dive Office for their help.



{\small
\bibliographystyle{IEEEtran}
\bibliography{refs}

@software{metashape,
  author = {{AgiSoft LLC}},
  title = {{Agisoft Metashape Professional} (Version 2.2.0)},
  url = {http://www.agisoft.com/downloads/installer/},
  year = {2025},
}

@inproceedings{kirillov2023segment,
  title={Segment anything},
  author={Kirillov, Alexander and Mintun, Eric and Ravi, Nikhila and Mao, Hanzi and Rolland, Chloe and Gustafson, Laura and Xiao, Tete and Whitehead, Spencer and Berg, Alexander C and Lo, Wan-Yen and others},
  booktitle={Proceedings of the IEEE/CVF international conference on computer vision},
  pages={4015--4026},
  year={2023}
}

@inproceedings{zheng2024coralscop,
  title={Coralscop: Segment any coral image on this planet},
  author={Zheng, Ziqiang and Liang, Haixin and Hua, Binh-Son and Wong, Yue Him and Ang, Put and Chui, Apple Pui Yi and Yeung, Sai-Kit},
  booktitle={Proceedings of the IEEE/CVF Conference on Computer Vision and Pattern Recognition},
  pages={28170--28180},
  year={2024}
}

@article{sauder2025coralscapes,
  title={The Coralscapes Dataset: Semantic Scene Understanding in Coral Reefs},
  author={Sauder, Jonathan and Domazetoski, Viktor and Banc-Prandi, Guilhem and Perna, Gabriela and Meibom, Anders and Tuia, Devis},
  journal={arXiv preprint arXiv:2503.20000},
  year={2025}
}

@inproceedings{beijbom2012automated,
  title={Automated annotation of coral reef survey images},
  author={Beijbom, Oscar and Edmunds, Peter J and Kline, David I and Mitchell, B Greg and Kriegman, David},
  booktitle={2012 IEEE conference on computer vision and pattern recognition},
  pages={1170--1177},
  year={2012},
  organization={IEEE}
}

@inproceedings{chen2021new,
  title={A new deep learning engine for CoralNet},
  author={Chen, Qimin and Beijbom, Oscar and Chan, Stephen and Bouwmeester, Jessica and Kriegman, David},
  booktitle={Proceedings of the IEEE/CVF international conference on computer vision},
  pages={3693--3702},
  year={2021}
}

@incollection{mahmood2017deep,
  title={Deep learning for coral classification},
  author={Mahmood, Ammar and Bennamoun, Mohammed and An, Senjian and Sohel, Ferdous and Boussaid, Farid and Hovey, Renae and Kendrick, Gary and Fisher, Robert B},
  booktitle={Handbook of neural computation},
  pages={383--401},
  year={2017},
  publisher={Elsevier}
}

@article{gonzalez2020monitoring,
  title={Monitoring of coral reefs using artificial intelligence: A feasible and cost-effective approach},
  author={Gonzalez-Rivero, Manuel and Beijbom, Oscar and Rodriguez-Ramirez, Alberto and Bryant, Dominic EP and Ganase, Anjani and Gonzalez-Marrero, Yeray and Herrera-Reveles, Ana and Kennedy, Emma V and Kim, Catherine JS and Lopez-Marcano, Sebastian and others},
  journal={Remote Sensing},
  volume={12},
  number={3},
  pages={489},
  year={2020},
  publisher={MDPI}
}

@article{hamilton2022unsupervised,
  title={Unsupervised semantic segmentation by distilling feature correspondences},
  author={Hamilton, Mark and Zhang, Zhoutong and Hariharan, Bharath and Snavely, Noah and Freeman, William T},
  journal={arXiv preprint arXiv:2203.08414},
  year={2022}
}

@inproceedings{radford2021learning,
  title={Learning transferable visual models from natural language supervision},
  author={Radford, Alec and Kim, Jong Wook and Hallacy, Chris and Ramesh, Aditya and Goh, Gabriel and Agarwal, Sandhini and Sastry, Girish and Askell, Amanda and Mishkin, Pamela and Clark, Jack and others},
  booktitle={International conference on machine learning},
  pages={8748--8763},
  year={2021},
  organization={PmLR}
}

@inproceedings{caron2021emerging,
  title={Emerging properties in self-supervised vision transformers},
  author={Caron, Mathilde and Touvron, Hugo and Misra, Ishan and J{\'e}gou, Herv{\'e} and Mairal, Julien and Bojanowski, Piotr and Joulin, Armand},
  booktitle={Proceedings of the IEEE/CVF international conference on computer vision},
  pages={9650--9660},
  year={2021}
}

@article{oquab2023dinov2,
  title={Dinov2: Learning robust visual features without supervision},
  author={Oquab, Maxime and Darcet, Timoth{\'e}e and Moutakanni, Th{\'e}o and Vo, Huy and Szafraniec, Marc and Khalidov, Vasil and Fernandez, Pierre and Haziza, Daniel and Massa, Francisco and El-Nouby, Alaaeldin and others},
  journal={arXiv preprint arXiv:2304.07193},
  year={2023}
}

@inproceedings{ayzenberg2024dinov2,
  title={Dinov2 based self supervised learning for few shot medical image segmentation},
  author={Ayzenberg, Lev and Giryes, Raja and Greenspan, Hayit},
  booktitle={2024 IEEE International Symposium on Biomedical Imaging (ISBI)},
  pages={1--5},
  year={2024},
  organization={IEEE}
}

@article{amir2021deep,
  title={Deep vit features as dense visual descriptors},
  author={Amir, Shir and Gandelsman, Yossi and Bagon, Shai and Dekel, Tali},
  journal={arXiv preprint arXiv:2112.05814},
  volume={2},
  number={3},
  pages={4},
  year={2021}
}

@article{docherty2024upsampling,
  title={Upsampling DINOv2 features for unsupervised vision tasks and weakly supervised materials segmentation},
  author={Docherty, Ronan and Vamvakeros, Antonis and Cooper, Samuel J},
  journal={arXiv preprint arXiv:2410.19836},
  year={2024}
}

@article{ong2025evaluation,
  title={An evaluation of a pre-trained transformer-based self-distillation model (DINOv2) for cross-domain plant species identification},
  author={Ong, Chin Ann and Tay, Fei Siang and Then, Yi Lung and McCarthy, Chris},
  journal={Neural Computing and Applications},
  pages={1--27},
  year={2025},
  publisher={Springer}
}

@article{shao2025multi,
  title={Multi-label classification for multi-temporal, multi-spatial coral reef condition monitoring using vision foundation model with adapter learning},
  author={Shao, Xinlei and Chen, Hongruixuan and Zhao, Fan and Magson, Kirsty and Chen, Jundong and Li, Peiran and Wang, Jiaqi and Sasaki, Jun},
  journal={arXiv preprint arXiv:2503.23012},
  year={2025}
}

@inproceedings{raine2024human,
  title={Human-in-the-loop segmentation of multi-species coral imagery},
  author={Raine, Scarlett and Marchant, Ross and Kusy, Brano and Maire, Frederic and Sunderhauf, Niko and Fischer, Tobias},
  booktitle={Proceedings of the IEEE/CVF Conference on Computer Vision and Pattern Recognition},
  pages={2723--2732},
  year={2024}
}

@inproceedings{todd2024adaptive,
  title={Adaptive multi-altitude search and sampling of sparsely distributed natural phenomena},
  author={Todd, Jessica E and McCammon, Seth and Girdhar, Yogesh and Roy, Nicholas and Yoerger, Dana},
  booktitle={2024 IEEE/RSJ International Conference on Intelligent Robots and Systems (IROS)},
  pages={8650--8657},
  year={2024},
  organization={IEEE}
}

@inproceedings{pizarro2004large,
  title={Large area 3D reconstructions from underwater surveys},
  author={Pizarro, Oscar and Eustice, Ryan and Singh, Hanumant},
  booktitle={Oceans' 04 MTS/IEEE Techno-Ocean'04 (IEEE Cat. No. 04CH37600)},
  volume={2},
  pages={678--687},
  year={2004},
  organization={IEEE}
}

@article{berild2024efficient,
  title={Efficient 3D real-time adaptive AUV sampling of a river plume front},
  author={Berild, Martin Outzen and Ge, Yaolin and Eidsvik, Jo and Fuglstad, Geir-Arne and Ellingsen, Ingrid},
  journal={Frontiers in Marine Science},
  volume={10},
  pages={1319719},
  year={2024},
  publisher={Frontiers Media SA}
}

@article{manderson2020vision,
  title={Vision-based goal-conditioned policies for underwater navigation in the presence of obstacles},
  author={Manderson, Travis and Higuera, Juan Camilo Gamboa and Wapnick, Stefan and Tremblay, Jean-Fran{\c{c}}ois and Shkurti, Florian and Meger, David and Dudek, Gregory},
  journal={arXiv preprint arXiv:2006.16235},
  year={2020}
}

@inproceedings{lin2024uivnav,
  title={Uivnav: Underwater information-driven vision-based navigation via imitation learning},
  author={Lin, Xiaomin and Karapetyan, Nare and Joshi, Kaustubh and Liu, Tianchen and Chopra, Nikhil and Yu, Miao and Tokekar, Pratap and Aloimonos, Yiannis},
  booktitle={2024 IEEE International Conference on Robotics and Automation (ICRA)},
  pages={5250--5256},
  year={2024},
  organization={IEEE}
}

@article{karapetyan2021human,
  title={Human diver-inspired visual navigation: Towards coverage path planning of shipwrecks},
  author={Karapetyan, Nare and Johnson, James V and Rekleitis, Ioannis},
  journal={Marine Technology Society Journal},
  volume={55},
  number={4},
  pages={24--32},
  year={2021},
  publisher={Marine Technology Society}
}

@article{girdhar2014autonomous,
  title={Autonomous adaptive exploration using realtime online spatiotemporal topic modeling},
  author={Girdhar, Yogesh and Giguere, Philippe and Dudek, Gregory},
  journal={The International Journal of Robotics Research},
  volume={33},
  number={4},
  pages={645--657},
  year={2014},
  publisher={SAGE Publications Sage UK: London, England}
}

@inproceedings{jamieson2020active,
  title={Active reward learning for co-robotic vision based exploration in bandwidth limited environments},
  author={Jamieson, Stewart and How, Jonathan P and Girdhar, Yogesh},
  booktitle={2020 IEEE International Conference on Robotics and Automation (ICRA)},
  pages={1806--1812},
  year={2020},
  organization={IEEE}
}

@article{flaspohler2019information,
  title={Information-guided robotic maximum seek-and-sample in partially observable continuous environments},
  author={Flaspohler, Genevieve and Preston, Victoria and Michel, Anna PM and Girdhar, Yogesh and Roy, Nicholas},
  journal={IEEE Robotics and Automation Letters},
  volume={4},
  number={4},
  pages={3782--3789},
  year={2019},
  publisher={IEEE}
}

@inproceedings{binney2010informative,
  title={Informative path planning for an autonomous underwater vehicle},
  author={Binney, Jonathan and Krause, Andreas and Sukhatme, Gaurav S},
  booktitle={2010 IEEE International Conference on Robotics and Automation},
  pages={4791--4796},
  year={2010},
  organization={IEEE}
}

@inproceedings{hagen2003hugin,
  title={The HUGIN 1000 autonomous underwater vehicle for military applications},
  author={Hagen, Per Espen and Storkersen, Nils and Vestgard, Karstein and Kartvedt, Per},
  booktitle={Oceans 2003. Celebrating the Past... Teaming Toward the Future (IEEE Cat. No. 03CH37492)},
  volume={2},
  pages={1141--1145},
  year={2003},
  organization={IEEE}
}

@inproceedings{jakuba2024exploring,
  title={Exploring the Aurora Vent Field: 4000 m Under Ice with the NUI Hybrid Remotely Operated Vehicle},
  author={Jakuba, Michael V and Curran, Molly and Naklicki, Victor and Isler, Tea and Klesh, Andrew and Buenger, H Jakob and Dalpe, Allisa and Lindzey, Laura and Silvia, Matthew and Loer, Rosemary and others},
  booktitle={2024 IEEE/OES Autonomous Underwater Vehicles Symposium (AUV)},
  pages={1--6},
  year={2024},
  organization={IEEE}
}

@article{choset2000coverage,
  title={Coverage of known spaces: The boustrophedon cellular decomposition},
  author={Choset, Howie},
  journal={Autonomous Robots},
  volume={9},
  number={3},
  pages={247--253},
  year={2000},
  publisher={Springer}
}
}

\end{document}